\pdfoutput=1

\documentclass[11pt]{article}

\usepackage{acl}
\usepackage{booktabs}
\usepackage{multirow}
\usepackage{enumitem}
\usepackage{graphics}
\usepackage{times}
\usepackage{latexsym}
\usepackage{graphicx}
\usepackage[T1]{fontenc}

\usepackage[utf8]{inputenc}

\usepackage{microtype}
\usepackage{amssymb}
\usepackage{amsmath}
\title{CINO: A Chinese Minority Pre-trained Language Model}

\author{
Ziqing Yang$^\dag$,
Zihang Xu$^\dag$,
Yiming Cui$^\ddag$$^\dag$\thanks{\ \ Email corresponding.},
Baoxin Wang$^\dag$,
Min Lin,\\
\textbf{Dayong Wu$^\dag$,
Zhigang Chen$^{\dag\S}$} \\
{$^\dag$State Key Laboratory of Cognitive Intelligence, iFLYTEK Research, Beijing, China} \\
{$^\ddag$Research Center for SCIR, Harbin Institute of Technology, Harbin, China} \\
{$^\S$Jilin Kexun Information Technology Co., Ltd., Changchun, China} \\
$^\dag$\tt\{zqyang5,zhxu13,ymcui,bxwang2,dywu2,zgchen\}@iflytek.com \\
$^\ddag$\tt ymcui@ir.hit.edu.cn}

\begin{document}
\maketitle
\begin{abstract}
Multilingual pre-trained language models have shown impressive performance on cross-lingual tasks. 
It greatly facilitates the applications of natural language processing on low-resource languages.
However, there are still some languages that the current multilingual models do not perform well on.
In this paper, we propose CINO (Chinese Minority Pre-trained Language Model), a multilingual pre-trained language model for Chinese minority languages.
It covers Standard Chinese, Yue Chinese, and six other ethnic minority languages.
To evaluate the cross-lingual ability of the multilingual model on ethnic minority languages, 
we collect documents from Wikipedia and news websites, and construct two text classification datasets, WCM (Wiki-Chinese-Minority) and CMNews (Chinese-Minority-News). 
We show that CINO notably outperforms the baselines on various classification tasks. The CINO model and the datasets are publicly available at \url{http://cino.hfl-rc.com}.

\end{abstract}

\section{Introduction}
The multilingual pre-trained language model (MPLM) is known for its ability to understand multiple languages, and its surprising zero-shot cross-lingual ability \cite{wu-dredze-2019-beto}. The zero-shot cross-lingual transfer ability enables the MPLM to be applied on the target languages with limited or even no annotated data by fine-tuning the MPLM on the source language with rich annotated data.
MPLMs greatly facilitate transferring the current NLP technologies to low-resource languages and reduce the cost of developing NLP applications for low-resource languages.

The existing public MPLMs such as mBERT \cite{devlin-etal-2019-bert}, XLM \cite{DBLP:conf/nips/ConneauL19} and XLM-R \cite{conneau-etal-2020-unsupervised} can handle 100 languages, but there are still some challenges on low-resource languages understanding:
\begin{itemize}[leftmargin=*]
  \item The size of pre-training corpora of some low-resource languages is small compared to the high-resource languages. This bias towards high-resource languages may harm the performance on low-resource languages.
  \item There are thousands of living languages in the world, but many languages have not been covered in the existing MPLMs, especially indigenous or ethnic minority languages.
  For example, Tibetan, a language spoken mainly by Tibetans around Tibetan Plateau, is absent from the CC-100 corpus. Therefore, the XLM-R tokenizer can not tokenize Tibetan scripts correctly, and XLM-R is not good at understanding Tibetan texts.
\end{itemize}

Recently, more advanced MPLMs have been proposed, such as ERNIE-M \cite{ouyang-etal-2021-ernie}, VECO \cite{luo-etal-2021-veco} and Unicoder \cite{huang-etal-2019-unicoder}. These models focus on multilingual training objectives, such as leveraging parallel sentences to improve the alignment between different languages, and have improved notably over XLM-R.
However, these models have not paid attention to the low-resource languages, so the problem remains unsolved.

For the above reasons, it is necessary to develop multilingual pre-trained language models for low-resource and ethnic minority languages.
In this paper, we focus on Chinese minority languages.
In China, Standard Chinese (Mandarin Chinese) is the predominant language. Besides Standard Chinese, we consider several most spoken minority languages. These languages are in different language families with varying writing systems, as summarized in Table \ref{minority_languages}.

\begin{table*}[h]
  \center
  \begin{tabular}{@{}llll@{}}
    \toprule
    \textbf{ISO Code} & \textbf{Language Name}  & \textbf{Language Family}  & \textbf{Writing System} \\ \midrule
    zh                & Standard Chinese (Mandarin)    & Sino-Tibetan              & Chinese characters      \\
    yue               & Yue Chinese (Cantonese)         & Sino-Tibetan                     & Chinese characters      \\
    bo                & Tibetan           & Sino-Tibetan                     & Tibetan script   \\
    mn                & Mongolian         & Mongolic                         & Traditional Mongolian script   \\
    ug                & Uyghur            & Turkic                           & Uyghur Arabic alphabet         \\
    kk                & Kazakh            & Turkic                           & Kazakh Arabic alphabet         \\
    za                & Zhuang            & Kra-Dai                          & Latin alphabet          \\
    ko                & Korean            & Isolate                          & Hangul                  \\
 \bottomrule
    \end{tabular}
  \caption{Families and writing systems of the languages covered by CINO.}\label{minority_languages}
\end{table*}

Although each of the listed minority languages is spoken by at least millions of people, their digital corpus resources are quite limited. For example, in the CC-100 corpus used by XLM-R, the size of the Uyghur (ug) corpus is 0.4 GB, which is about 1\% of the Chinese (Simplified) corpus (46.9 GB); also, there are no Tibetan (bo) or (traditional) Mongolian (mn) corpora in the CC-100.

We propose a multilingual pre-trained language model named CINO (\textbf{C}hinese M\textbf{ino}rity Pre-trained Language Model), which covers Standard Chinese, Yue Chinese (Cantonese) and six ethnic minority languages. As far as we know, this is the first multilingual pre-trained language model for the Chinese minority languages. CINO largely has the same structure as XLM-R and has been adapted for minority languages by resizing its vocabulary and adopting a fast masked language modeling objective for the pre-training.

The reason for training a multilingual pre-trained model rather than multiple monolingual pre-trained models is threefold. First, a multilingual model is more convenient than multiple monolingual models. 
Second, for low-resource languages, multilingual pre-training leads to better performance than monolingual pre-training \cite{conneau-etal-2020-unsupervised,wu-dredze-2020-languages}.
Third, a multilingual pre-trained model provides cross-lingual transfer ability, which reduces the data annotation cost for low-resource languages. Studies have also shown that pre-training with more languages leads to better cross-lingual performance on low-resource languages \cite{conneau-etal-2020-unsupervised}.

The public natural language understanding tasks in Chinese minority languages are extremely limited. In this work, we construct two multilingual datasets from two data sources to support evaluating the zero-shot cross-lingual ability of MPLMs on the Chinese minority languages: (1)
The \textbf{WCM} (\textbf{W}iki-\textbf{C}hinese-\textbf{M}inority) dataset is a multilingual text classification dataset built from Wikipedia corpora, with 10 classes, consisting of 63k examples. (2) \textbf{CMNews} (\textbf{C}hinese \textbf{M}inority \textbf{N}ews) dataset is a multilingual news classification dataset with 8 classes, built from the crawled news and the pre-existing news datasets, consisting of 57k examples.

To evaluate CINO from different perspectives, we run experiments on Tibetan News Classification Corpus (TNCC), Korean news topic classification (YNAT), WCM, and CMNews. 
Results show that CINO has acquired the ability of minority language understanding and outperforms the existing baselines on the Chinese minority languages. 

To summarize, our contributions are:
\begin{itemize}[topsep=0pt,itemsep=0pt]
  \item We introduce CINO, the first multilingual pre-trained language model for Chinese minority languages. Besides Standard Chinese, CINO covers Yue Chinese and six ethnic minority languages.
  \item We construct two multilingual text classification datasets for Chinese minority languages. They are used for evaluating the cross-lingual and multilingual abilities of the ethnic minority language model.
  \item Experiments show that CINO achieves notable improvements over the baselines. Furthermore, by making the model public, CINO will be a useful resource on Chinese minority languages and facilitate related research.
 \end{itemize}

\section{Related Work}
\subsection{Pre-trained Language Models}
\textbf{Multilingual Pre-trained Language Models}.
\citet{devlin-etal-2019-bert} introduced the first multilingual pre-trained language model mBERT trained with Masked Language Modeling (MLM). \citet{DBLP:conf/nips/ConneauL19} proposed Translation Language Modeling (TLM) to train the multilingual model with cross-lingual supervision. Since then, various kinds of multilingual pre-training objectives have been proposed.
Unicoder \cite{huang-etal-2019-unicoder} trains the model with the objectives including cross-lingual word recovery, cross-lingual paraphrase classification and cross-lingual MLM. 
InfoXLM \cite{chi-etal-2021-infoxlm} proposed a pre-training task based on contrastive learning from an information-theoretic perspective.
\citet{pan-etal-2021-multilingual} also introduced an alignment method based on contrastive learning.
\citet{DBLP:conf/iclr/CaoKK20} proposed an explicit word-level alignment procedure. 
ERNIE-M \cite{ouyang-etal-2021-ernie} integrates back-translation into the pre-training process.
VECO \cite{luo-etal-2021-veco} uses a cross-attention module to build the interdependence between languages explicitly.
In this work, we only use non-parallel data and an objective similar to MLM for pre-training CINO.

\textbf{Non-English Pre-trained Language Models and Benchmarks}. 
Many pre-trained models have been trained on English corpora, or corpora that are heavily biased toward English. To make NLP techniques accessible to people from different cultures, researchers have developed pre-trained models and benchmarks targeting different languages: FlauBERT and the FLUE benchmark for French \cite{le2020flaubert}, KLUE-BERT and the KLUE benchmark for Korean \cite{park2021klue}, IndoBERT and the IndoLEM benchmark for Indonesian \cite{koto-etal-2020-indolem}, and there are Chinese-BERT-wwm \cite{cui-etal-2021-pretrain} and Arabic BERT AraBERT \cite{antoun2020arabert}. However, there are no pre-trained language models targeting Chinese ethnic minority languages.

\subsection{Language Diversity in China}
There are 56 ethnic groups and more than 80 languages in China. Standard Chinese (Mandarin) is the official language, spoken mainly by ethnic Han Chinese, which accounts for more than 90\% of the total population. Ethnic minorities have their own languages. According to the study in \citet{moseley2010atlas}, the ethnic minority languages Mongolian, Uyghur, Kazakh, Tibetan,Yi, and Korean are safe (five of them are covered by CINO), which are spoken by about 25 million people, while the rest are in unsafe or endangered status.  

Besides the ethnic minority languages, there are dialects and varieties of Chinese across the country. In this work, we consider Yue Chinese (also known as Cantonese), a widely used group of varieties of Chinese in Southern China and have been carried by immigrants to Southeast Asia and many other parts of the world.

Some languages in Table \ref{minority_languages} are spoken and widely used in more than one country, such as Korean, Mongolian and Kazakh. In this work, we named them as \emph{minority} languages based on their status in China.

\section{CINO Model}

In this section, we present the CINO model structure and the pre-training methodology. We denote by $N$ the number of pre-training languages, $\mathcal{C}_i$ the monolingual corpus of the $i$th language ($i=1,\ldots,N$). Let $n_i$ be the number of sentences and $l_i$ be the mean sequence length in $\mathcal{C}_i$. Let $c_i$ represent the total number of tokens of $\mathcal{C}_i$.
\subsection{Model Structure}\label{sec:model_struct}

\begin{figure}[t]
 \includegraphics [width=\linewidth] {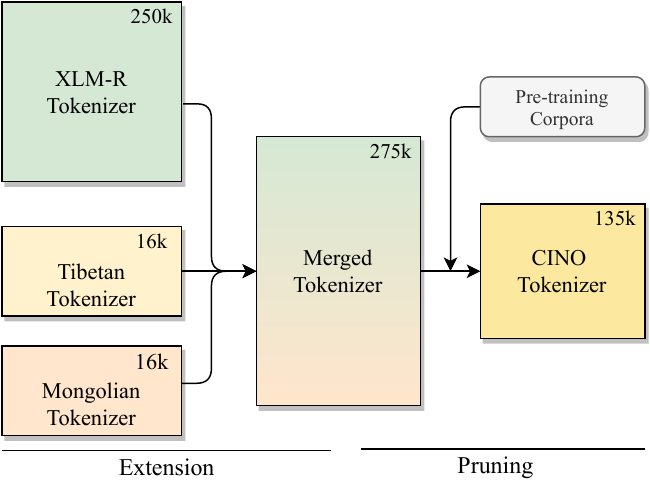}
  \caption{We extend the XLM-R tokenizer with a Tibetan tokenizer and a Mongolian tokenizer, then remove the redundant tokens to obtain the CINO tokenizer.}\label{fig:cino-tokenizer}
\end{figure}

CINO is a multilingual transformer-based model with the same architecture as XLM-R. For the CINO-base, it has 12 layers, 768 hidden states, and 12 attention heads; for the CINO-large, it has 24 layers, 1024 hidden states, and 16 attention heads. The main differences between CINO and XLM-R are the word embeddings and the tokenizer. We start from the word embeddings and the tokenizer of XLM-R and adapt them for the minority languages by vocabulary extension and vocabulary pruning, as depicted in Figure \ref{fig:cino-tokenizer}.

\textbf{Vocabulary Extension}. The original XLM-R tokenizer does not recognize Tibetan scripts and Traditional Mongolian scripts, so we extend the XLM-R tokenizer and XLM-R word embeddings matrix with additional tokens. 

We train sentence-piece tokenizers for Tibetan and Mongolian on their monolingual pre-training corpora respectively. Each of the tokenizers has a vocabulary size of 16,000. Then we merge the vocabularies from the Tibetan and Mongolian tokenizers into the original XLM-R tokenizer. The merged tokenizer has a vocabulary size of 274,701.

To extend the word embeddings, we resize the original word embeddings matrix of shape $V\times D$ to $V'\times D$ by appending new rows, where $D$ is the hidden size, $V$ is the original vocabulary size, $V'$ is the new vocabulary size. The new rows represent the word vectors of the new tokens from the merged tokenizer. They are initialized with a Gaussian distribution of mean 0.0 and variance 0.02.

\textbf{Vocabulary Pruning}. Next, we prune the word embeddings matrix to reduce the model size. 
We tokenize the pre-training corpora with the merged tokenizer, and remove all the tokens that have not appeared in the corpora from the merged tokenizer's vocabulary and the word embeddings matrix. The above process discards 139,342 tokens.

Finally, we obtain the CINO model structure with a vocabulary size of 135,359, a model size of 728 MB for the base model, 1.7 GB for the large model, 68\% and 79\% size of XLM-R-base and XLM-R-large, respectively. A smaller vocabulary size leads to not only a memory-friendly model but also a faster model by reducing the cost of computing the log-softmax in the MLM task. The time cost of each iteration in pre-training is reduced by approximately 35\% by reducing the vocabulary size from 270k to 140k.

\subsection{Pre-training}\label{sec:pre-training}
We adopt the MLM objective for pre-training. In addition, we apply the following strategies for balancing training data and faster pre-training.

\subsubsection{Resampling Strategy}
To balance the data size between high-resource and low-resource languages, \citet{DBLP:conf/nips/ConneauL19} and \citet{chi-etal-2021-infoxlm} have applied a multinomial sampling strategy. An example in the $i$th language is sampled with the probability 
\begin{equation}
	p_i=\frac{n_i^\alpha}{\sum_k^N n_k^\alpha},\label{sampling_ratio1}
\end{equation} where $\alpha\in (0,1]$ is a hyperparameter.

However, if the mean sequence lengths of different corpora are different, it may lead to an undesired data bias.\footnote{In most cases, we could join short sequences to form long sequences of a uniform length. But some corpora we use consist of short sentences. Joining them as a long sequence leads to semantically incoherence.} To see this, we use $\tilde c_i$ to denote the number of tokens seen during training. We have $\tilde c_i \propto p_i l_i$ and $\tilde c_i = K c_i$ for all $i=1\ldots N$ if $\alpha=1$. $K$ is a constant that only depends on the number of training steps. 
If two languages $i$ and $j$ that have the same number of tokens, i.e., $c_i=c_j$, but with $n_i > n_j$ and $l_i < l_j$. With the sampling ratio in \eqref{sampling_ratio1}, we get $\tilde c_i < \tilde c_j$ if $\alpha < 1$ although the original corpora are of the same size. To remedy this, we introduce the dependence on the mean sequence length $l_i$. The sampling probability is
\begin{equation}\label{sampling_ratio2}
  p_i = \frac{n_i^\alpha / l_i^\beta}{\sum_k^N n_k^\alpha / l_k^\beta},
\end{equation}
where $\beta\in [0,1]$. Setting $\beta = 1 - \alpha$, the number of training tokens in the $i$th language is 
\begin{equation}
	\tilde c_i \propto 	p_i l_i \propto n_i^\alpha l_i^{1-\beta} = (n_i l_i)^\alpha = c_i^\alpha.
\end{equation}
Therefore, corpora of equal size will be trained with an equal number of tokens.

\begin{figure}[t]
 \includegraphics [width=\linewidth] {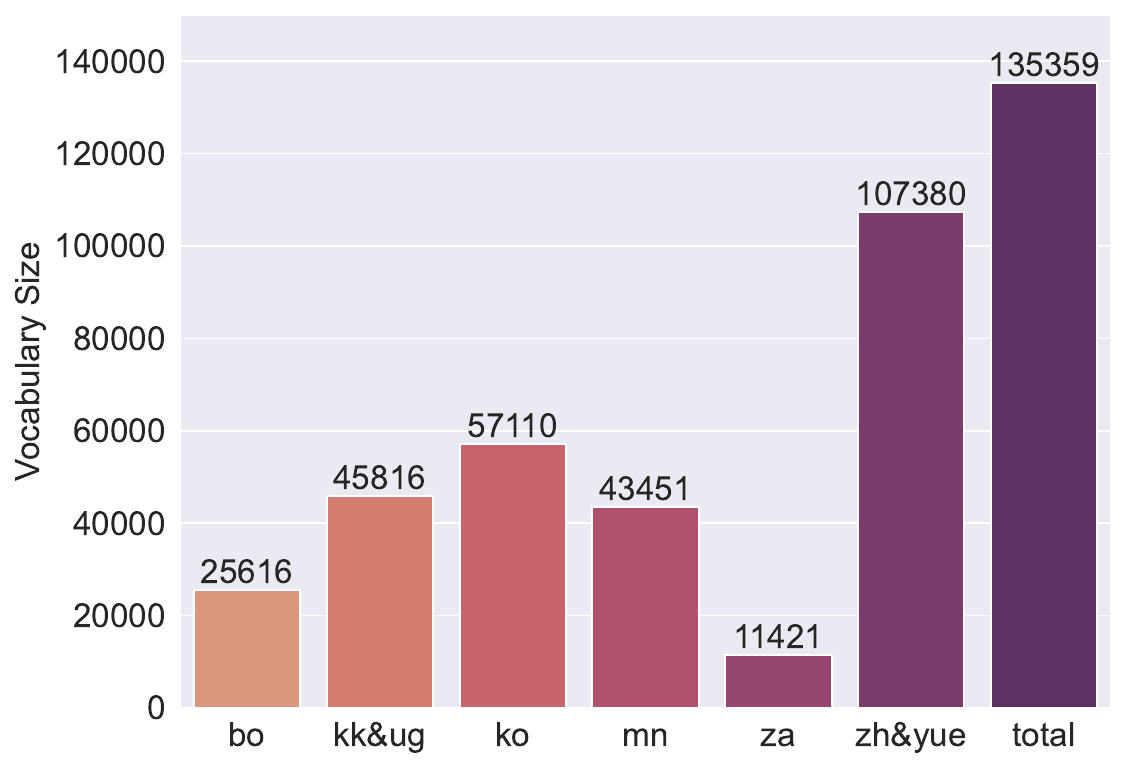}
  \caption{The vocabulary size counted from the corpus of each language. We merge the vocabularies of the languages that have similar writing systems.}\label{fig:tokens}
\end{figure}

\subsubsection{Fast Masking Language Modeling}

Table \ref{minority_languages} shows that the languages we consider have distinguished writing systems, which implies that
the vocabulary of each language only takes up a fraction of the whole vocabulary, as shown in Figure \ref{fig:tokens}.
By taking advantage of this fact, the computational costs can be reduced if the model only makes MLM predictions over the vocabulary of the specific language of the input examples rather than the whole vocabulary.

Suppose the example is in the $i$th language.
We denote by $\mathcal{V}$ the full vocabulary, and $\mathcal{V}_i\subset \mathcal{V}$ the vocabulary of the $i$th language, which is obtained by tokenizing the $i$th language's monolingual corpus. 
Let $(c,x)$ denote the input text sequence, where $x$ is the masked token, and $c$ is the context.
By limiting the prediction of the masked token to $\mathcal{V}_i$, the MLM loss of the masked token $x$ is 
\begin{equation}\label{mlm_loss}
\mathcal{L}^{(i)}_{\textrm{MLM}} = -\log \frac{\exp(g(c)\cdot E(x))}{\sum_{x'\in\mathcal{V}_i} \exp{(g(c)\cdot E(x'))}},
\end{equation}
where $g(\cdot)$ is the transformer encoder and $E(\cdot)$ is the look-up operation that returns the embeddings. 

In order to calculate the loss \eqref{mlm_loss} efficiently, during training, we group examples by language so that each batch contains examples in a single language.

With the objective \eqref{mlm_loss} for pre-training, we have observed 10\% time reduction and no significant performance drop compared to the original MLM objective, which predicts over the whole vocabulary. Combined with the speedup by vocabulary pruning, the pre-training time cost is reduced by about 40\% in total.

\section{Text Classification Datasets for Minority Languages}

Multilingual tasks have been used widely to evaluate the cross-lingual transferability of multilingual models \cite{hu2020xtreme}. Nevertheless, the pre-existing multilingual datasets hardly cover the Chinese ethnic minority languages. For example, Tibetan, Mongolian and Uyghur have never appeared in any task in the XTREME benchmark.
To evaluate the cross-lingual transferability of CINO, we construct two text classification datasets \textbf{WCM} (\textbf{W}ikipedia-\textbf{C}hinese-\textbf{M}inority) and \textbf{CMNews} (\textbf{C}hinese-\textbf{M}inority-\textbf{N}ews).
\subsection{WCM Dataset}

\textbf{Data Collection and Annotation}.
WCM is based on the data from Wikipedia. It covers seven languages: Mongolian, Tibetan, Uyghur, Kazakh, Korean, Cantonese, and Standard Chinese. 
We build the dataset from the Wikipedia page dumps and the Wikipedia category dumps\footnote{https://dumps.wikimedia.org/other} of the languages in question.

To annotate the data, we first generate a category graph for each language. Each node represents a category, and each edge stands for the affiliation between a pair of categories. By referring to the category system of Chinese Wikipedia, we choose ten categories for the classification task: Art, Geography, History, Nature, Science, Personage, Technology, Education, Economy, and Health. Then, we start from the categories of each page and backtrack along the routes in the category graph until reaching one of the ten target categories, and we set this category as the label of that page. Owing to some affiliation conflicts, like one subcategory belonging to two categories simultaneously, we reconstructed the graph by removing certain edges between the 10 target categories and their subcategories which are assessed as unreasonable by our human evaluation team.

\textbf{Data Cleaning}.
After getting the labeled data, we apply several strategies to improve the quality of the datasets. We remove dirty data like large blocks of URLs and file paths. Then, the examples are filtered by their lengths (after being tokenized by the CINO tokenizer) by removing those examples shorter than 20 or longer than 1024 tokens.\footnote {We discard examples that are too long because long examples likely cover multiple topics while we assign a single label to each example.}

\textbf{Subsampling}. Since there are both high-resource languages like Korean and low-resource languages like Uyghur, we down-sample the data in the high-resource languages and the high-resource categories to balance the numbers of examples among different languages and different categories. We fix the size of the training set (Chinese articles) to 32K and downsample the datasets of the languages with abundant articles to about $5\%\sim 20\%$ size of the training set. Similarly, we also downsampled some categories if they dominate in some languages. We did not apply the above process to Uyghur due to its extreme scarcity.

Finally, we obtain 63,137 examples. WCM contains the train/dev/test set for Standard Chinese and only test sets for other languages.
The detailed distribution is listed in Appendix \ref{sec:stat-data}.

\begin{table*}[htbp]
\small
\center
\begin{tabular}{@{}l|lcccccccc@{}}
\toprule
\textbf{Dataset}                 & \textbf{}    & \textbf{mn} & \textbf{bo} & \textbf{ug} & \textbf{kk} & \textbf{ko} & \textbf{yue} & \textbf{zh} & \textbf{Total} \\ \midrule
\multirow{3}{*}{\textbf{WCM}}    & \textbf{\# Samples}   & 27          & 5           & 4           & 52          & 43          & 49           & 20          & 200            \\
                                 & \textbf{\# Correctly Labeled}      & 24          & 4           & 4           & 49          & 34          & 43           & 19          & 177            \\
                                 & \textbf{Matching Acc} & 88.9\%      & 80\%        & 100\%       & 94.2\%      & 79.1\%      & 87.8\%       & 95.0\%      & 88.5\%         \\ \midrule
\multirow{3}{*}{\textbf{CMNews}} & \textbf{\# Samples}  & 11          & 34          & 24          & 14          & 10          & 23           & 84          & 200            \\
                                 & \textbf{\# Correctly Labeled} & 8           & 31          & 24          & 14          & 10          & 20           & 80          & 187            \\
                                 & \textbf{Matching Acc} & 72.7\%      & 91.2\%      & 100\%       & 100\%       & 100\%       & 87.0\%       & 95.2\%      & 93.5\%         \\ \bottomrule
\end{tabular}
\caption{Results of human evaluation of the sampled examples from WCM and CMNews.}
\label{table_human_evaluation}
\end{table*}

\subsection{CMNews Dataset}

\textbf{Data Collection and Annotation}. To collect the minority language examples, we crawl the news from the news websites in ethnic minority languages and record the category to which each news item belongs.
To collect the Chinese news, we reuse the pre-existing dataset SogouCS News \cite{DBLP:conf/www/WangZMR08} and CAIL 2018 \cite{DBLP:journals/corr/abs-1807-02478}. We select the appropriate categories and down-sample the two datasets to make the whole dataset more balanced.

After gathering the raw data from all the languages, we first merge the categories that have similar meanings (for example, we merge the categories \textit{Finance} and \textit{Economy}). Since the definition of news category may vary from website to website and language to language,  we remove the categories that are not consistent in different languages by manually checking a sampled subset. We also remove the categories that do not appear in more than two languages. Finally, we obtain a dataset containing eight categories: Education, Sports, Health, Tourism, Legal, Economy, Culture, and Society.

\textbf{Data Cleaning}. The crawled news is much cleaner than the Wikipedia pages, and each document naturally belongs to only one category. Therefore we only perform length filtering by keeping the documents that contain more than 30 tokens after tokenization. 

The dataset contains 56,764 examples in total. 
We split the dataset into a training set and a development set. The detailed distribution is listed in Appendix \ref{sec:stat-data}.

\subsection{Human Evaluation}
To assess the quality of the datasets, we randomly sample 200 examples from WCM and 200 examples from CMNews and manually check whether the contents of the examples match their labels. The results are shown in Table \ref{table_human_evaluation}. \textbf{Matching Acc} denotes how many examples match their labels under human evaluation. We find that $88.5\%$ of the sampled examples from WCM and $93.5\%$ of the sampled examples from CMNews are correctly labeled, which shows CMNews has less noise.

\section{Experiments} 
\subsection{Pre-training Setup}

\textbf{Pre-training Data}. 
We randomly sample a subset dataset from the public base version of WuDaoCorpora \cite{YUAN202165} as the Standard Chinese corpus; the corpora of the minority languages are in-house data, consisting of short monolingual sentences.
 The total corpora size is 28 GB. The statistics of the pre-training corpora are listed in Appendix \ref{sec:stat-corpora}. 

\textbf{Experiment Settings}.
CINO is trained with the fast MLM objective \eqref{mlm_loss} with the masking probability is 0.2 and the max sequence length 256. 
We initialize the parameters of CINO with XLM-R. We use the AdamW optimizer \cite{adamW} with the peak learning rate of 2e-4 for the base model and 1e-4 for the large model. 
The learning rate is scheduled with 10k and 5k warmup steps followed by a linear decay for the base and the large model respectively.
The sampling hyperparameter $\alpha$ is set to $0.7$.
We train the model with the batch size of 4,096 for 150k steps for the base model, and the batch size of 8,192 for 75k steps for the large model. The pre-training is performed on 16 NVIDIA A100 GPUs. The full pre-training hyperparameters are summarized in Appendix \ref{sec:pt-hp}.

\subsection{Downstream Evaluation}

How does CINO perform on the newly introduced languages? How does CINO perform on the languages pre-existing in XLM-R? Does CINO show multilingual and cross-lingual abilities? To answer these questions, we evaluate CINO on (1) Tibetan News Classification Corpus \cite{DBLP:conf/cncl/QunLQH17} (\textbf{TNCC}); (2) Korean news topic classification \cite{park2021klue} (\textbf{YNAT}); (3) \textbf{WCM} and \textbf{CMNews}.
On TNCC and YNAT,\footnote{The splitting sizes of TNCC and YNAT are listed in Appendix \ref{sec:stat-data}.} we evaluate the in-language model performance, i.e., we train and evaluate the model on the same language.
On WCM and CMNews, we evaluate the cross-lingual ability. We describe the details in Section \ref{sec:r_and_d}.

For each task and each model, we run the experiment five times with different seeds and report the mean metrics. The fine-tuning hyperparameters of each experiment are listed in Appendix \ref{sec:ft-hp}.

\subsection{Baselines}
Besides the common multilingual pre-trained models \textbf{mBERT} and \textbf{XLM-R},
we compare CINO models with the following baselines on some tasks.

\noindent \textbf{XLM-R-Ext}. We extend and prune the vocabulary of XLM-R as described in Section \ref{sec:model_struct}. This model is the un-pretrained CINO. The embeddings of Tibetan and Mongolian are randomly initialized, and the other parameters are the same as XLM-R.

\noindent \textbf{KLUE-BERT-base}. This is a Korean pre-trained model proposed in \citet{park2021klue}. Although KLUE-BERT-base is a base-sized model, it outperforms other large models on the YNAT task except for XLM-R-large.

\noindent \textbf{TextCNN} is a simple and light-weight model for text classification tasks \cite{textCNN}. The word embedding dimension is set to 300. After the embedding layer, we apply three convolution layers in parallel with the number of out-channels 100, kernel size 3,4, and 5, respectively. Finally, we concatenate the outputs from the convolution layers and apply a two-layer fully-connected network with ReLU activation to perform the classification. We train the TextCNN from scratch with randomly initialized model parameters and word embeddings.

\noindent \textbf{Word2vec (Tibetan)}. We first train the word embeddings using word2vec \cite{DBLP:journals/corr/abs-1301-3781, DBLP:journals/corr/MikolovSCCD13} on the TNCC training set. The embedding dimension is set to 300. To perform the classification task, we average the word embeddings of each sample, then feed the results to a trainable linear layer that outputs the logits.

\subsection{Results and Discussions}\label{sec:r_and_d}

\begin{table}[t]
  \center
  \small
  \begin{tabular}{@{}lcc|cc@{}}
  \toprule
  \multirow{2}{*}{\textbf{Model}}  & \multicolumn{2}{c}{\textbf{TNCC Dev}}  & \multicolumn{2}{c}{\textbf{TNCC Test} }        \\ \cmidrule(l){2-5} 
  & Acc      & Macro-F1        & Acc      & Macro-F1 \\ \midrule
  TextCNN & 69.4  & 65.7 & 62.8 & 66.6 \\ 
Word2vec (Tibetan) & 70.1  & 67.7 & 70.2 & 68.0 \\ \midrule
\textit{base models}      
              \\
              mBERT & 22.9 & 4.8 & 22.8 & 5.5 \\
              mBERT (p.t.) & 63.9 & 56.2 &  61.8 & 56.4 \\
  XLM-R-base      & 35.1    & 20.2  & 31.1 & 21.1 \\
  XLM-R-base (p.t.)     & 34.2    & 21.5  & 31.4 & 19.9 \\
  XLM-R-Ext-base & 55.7 & 43.2 & 55.0 & 42.1 \\
  CINO-base     & 74.8 & 71.4  & 73.1 & 70.0 \\ \midrule
\textit{large models}                    \\
  XLM-R-large           & 35.7   & 26.4         & 32.8 & 27.3                     \\
    {XLM-R-Ext-large} &  31.6 & 13.0 & 29.2 &  12.2 \\
  CINO-large           & \textbf{76.3} & \textbf{73.7}      &  \textbf{75.4} & \textbf{72.9}     \\       \bottomrule       
  \end{tabular}
  \caption{Model performance on the Dev and Test sets of Tibetan text classification task TNCC. \textit{p.t.} is short for \textit{pre-tokenized}.}\label{results1}
  \end{table}

  \begin{table}[t!]
  \center
  \small
  \begin{tabular}{@{}lcc@{}}
  \toprule
  \multirow{2}{*}{\textbf{Model}}  & \multicolumn{2}{c}{\textbf{YNAT Dev} }     \\ \cmidrule(l){2-3} 
                              &       Acc      & Macro-F1              \\ \midrule
  mBERT \cite{park2021klue} & - & 82.6${}^{\dagger}$ \\ 
  XLM-R-base \cite{park2021klue} & - & 84.5${}^{\dagger}$ \\ 
  XLM-R-large \cite{park2021klue} & - & 87.3${}^{\dagger}$ \\
  KLUE-RoBERTa-large \cite{park2021klue} & - & 85.9${}^{\dagger}$ \\
  KLUE-BERT-base \cite{park2021klue} & - & 87.0${}^{\dagger}$ \\
  \midrule
\textit{base models}                    \\
 mBERT & 82.9 & 82.8 \\
  XLM-R-base      & {85.1}    & {85.0}  \\
  KLUE-BERT-base & 87.0 & \textbf{87.1} \\
  CINO-base     & {86.1} & {85.9}  \\ \midrule
\textit{large models}                    \\
  XLM-R-large           & 87.0   & 86.8           \\
  CINO-large           & \textbf{87.3} & 87.0   \\       \bottomrule       
  \end{tabular}
  \caption{Model performance on the Dev set of Korean text classification task YNAT. The results marked with ${}^{\dagger}$ are taken from the KLUE paper \cite{park2021klue}. The rest results are from our experiments.}\label{results-ynat}
  \end{table}

\begin{table*}[t!]
\small
  \center
  \begin{tabular}{@{}c|lccccccccc@{}}
  \toprule
    \multirow{7}{*}{\begin{tabular}[c]{@{}c@{}}\textbf{WCM}\\ \\ \textit{zh} $\rightarrow$ \textit{min}.\end{tabular}} &
\textbf{Model} & \textbf{bo} & \textbf{kk} & \textbf{ko} & \textbf{mn} & \textbf{ug} & \textbf{yue}  & \textbf{zh} & \textbf{Avg (Minorities)} &\textbf{Avg (All)} \\ \cmidrule(l){2-11}
& \textit{base models}   \\
 & {XLM-R-base}    & 19.0   & 16.7   & 43.2   & 15.2   & 23.3  & 58.3   & 78.1 & 29.3   & 36.2   \\
 & {CINO-base}     & 36.2   & 43.2   & \textbf{44.9}  &  39.1   & \textbf{33.4}  & 59.7   & 78.0 & 42.6   & 47.6  \\ \cmidrule(l){2-11}
&  \textit{large models}   \\
 & {XLM-R-large}    & 18.4   & 32.9   & 43.8   & 22.2   & 27.8   & \textbf{60.0}    & 77.3  & 34.2   & 40.3   \\
 & {CINO-large}     & \textbf{40.6}   & \textbf{44.8}   & \textbf{44.8}  & \textbf{41.6}     & 28.8  & {59.8}  & \textbf{79.2} & \textbf{43.3}   &  \textbf{48.4}  \\ 
  \midrule \midrule
    \multirow{7}{*}{\begin{tabular}[c]{@{}c@{}}\textbf{CMNews}\\ \\ \textit{min}. $\rightarrow$ \textit{zh}\end{tabular}} &
\textbf{Model} & \textbf{bo} & \textbf{kk} & \textbf{ko} & \textbf{mn} & \textbf{ug} & \textbf{yue}  & \textbf{zh} & \textbf{Avg (Minorities)} &\textbf{Avg (All)} \\ \cmidrule(l){2-11}
& \textit{base models}   \\
 & {XLM-R-base}    & 38.1	&69.6	&88.3	&35.1	&77.5 (67.7/88.6)	&\textbf{87.8}	&58.6	&66.1	&65.0   \\
 & {CINO-base}     &85.5	 &79.2	&89.0	&77.3	&77.4 (77.0/78.0)	&86.9	&68.8 &82.6	&80.6 \\ \cmidrule(l){2-11}
&  \textit{large models}   \\
&  {XLM-R-large} & 30.1	&80.8	&88.9	&30.8 &	\textbf{85.1} (76.4/91.0)  &87.5	&63.6 &	67.2	&66.7  \\
&  {CINO-large}  &\textbf{86.8}	&\textbf{83.0}	&\textbf{90.3}	&\textbf{79.4}	&{78.8} (68.4/91.3)	&\textbf{87.9}	&\textbf{71.2}	&\textbf{84.4}	&\textbf{82.5}    \\ \bottomrule
  \end{tabular}
  \caption{Model performance on the WCM and CMNews. The metric on each language is macro-F1. \textbf{Avg (Minorities)} is the mean score over languages other than zh; \textbf{Avg (All)} is the mean score over all languages. We bold any score within 0.1 of the best on each language. The results in the parentheses are the min and the max values of five runs.}\label{results_WCM}
\end{table*}

\subsubsection{TNCC}
\textbf{How does CINO perform on the newly introduced language?} We evaluate CINO on TNCC, a Tibetan classification dataset with 12 classes. The original work \citep{DBLP:conf/cncl/QunLQH17} proposes a news title classification and a news document classification. Here we conduct the news document classification only. The task is to predict the topic of each document. Because there are no official splits available, we split the dataset into a training set, a development set and a test set with a ratio of 8:1:1. Since the texts in the dataset have been pre-tokenized (spaces have been added between words), we remove the spaces between words and tokenize the texts with the pre-trained tokenizer unless otherwise specified.
We select the best checkpoint based on its macro-F1 score. We also report the accuracy score for reference. 

The results are listed in Table \ref{results1}. Compared among the pre-trained models, {XLM-R} series have low scores since the vocabulary is not adapted for the Tibetan language and has not been pre-trained on the Tibetan corpus. While {XLM-R-Ext-base} has an extended vocabulary and significantly outperforms {XLM-R-base} even without being pre-trained on the target language. Finally, by pre-training on the minority languages corpora, {CINO} is adapted to the new language and outperforms {XLM-R} and {XLM-R-Ext} notably.

{mBERT} achieves better results when fine-tuned on the pre-tokenized data (but there are still many tokens being mapped to \texttt{[UNK]}). Due to the difference in the tokenization algorithms used by mBERT and XLM-R, XLM-R does not benefit from using pre-tokenized data.

 TextCNN and Word2vec (Tibetan) surprisingly achieve competitive scores and outperforms {XLM-R-Ext-base}. It is possibly due to the difficulty in the optimization of large models such as XLM-R with limited training data. As we continue increasing the model size, the performance gets worse, as can be seen from comparing the scores of {XLM-R-base-Ext} and {XLM-R-large-Ext}.

\subsubsection{YNAT}
\textbf{How does CINO perform on the minority languages pre-existing in XLM-R?}
We evaluate CINO on YNAT, a Korean text classification dataset with 7 classes. We select the best checkpoint based on its macro-F1 score. The results are listed in Table \ref{results-ynat}. 
{CINO-base} outperforms {XLM-R-base}, while {CINO-large} is better than {XLM-R-large} by our reimplementation but lower than the score reported in \citet{park2021klue}. CINO-large is also comparable to {KLUE-BERT-base}. 
 
Notice that Korean is not a low-resource language in XLM-R (the size of the Korean corpus is 54 GB in the CC-100), thus XLM-R may have learned Korean well. To significantly outperform XLM-R and KLUE-BERT-base, we expect that longer training time and more data are required.

\subsubsection{WCM and CMNews}
\textbf{Does CINO show multilingual and cross-lingual abilities?}
We use these two datasets to evaluate the cross-lingual and multilingual abilities.
We take macro-F1 as the metric on each language, and the \textit{Avg}  is the arithmetic mean of the macro-F1 scores.

On the WCM dataset, we train models on the Chinese training set and test it on all the languages, so the results show how well the model transfers the knowledge from Chinese to the minority languages; the best checkpoint of each run is selected based on its score on Chinese;
On the CMNews dataset, we train models on the minority languages and  the Chinese data is zero-shot; the best checkpoint is selected based on its score on minority languages.
The results are listed in Table \ref{results_WCM}.

On WCM, \textbf{Avg (Minorities)} score shows that CINO has superior zero-shot performance over XLM-R. By inspecting the detailed performance on each language, we see that CINO most significantly outperforms XLM-R on Tibetan,  Kazakh, Mongolian and Uyghur, which have been insufficiently pre-trained in XLM-R.

On CMNews, because CINO has been adapted to minority languages, it learns more effectively than XLM-R by leveraging the examples in all the languages. \textbf{zh} score shows that CINO transfers better than XLM-R. CINO also outperforms XLM-R on almost all the minority languages except for \textbf{ug}, where there is a large gap. To find out the reason, we list the min and the max \textbf{ug} scores of five runs. We see that there is a large variance. CINO-large achieves the highest score among all runs, but its average score is lower than XLM-R-large. The unstable performance may be the main reason that explains the gap.

\section{Discussion on Limitations}

\textbf{Coverage of ethnic minority languages}. Due to the scarcity of minority language corpora, CINO only covers Standard Chinese and some of the most popular minority languages and dialects. While being spoken by millions of people, some languages, such as the Yi language,  are omitted in this study since we can not find sufficient data for pre-training.

\textbf{Pre-training objectives}. In our early trials of multilingual pre-training, we leveraged both monolingual and bilingual parallel data, and combined the MLM objective with a cross-lingual alignment objective, similar to the TLM objective used in \citet{chi-etal-2021-infoxlm} and \citet{DBLP:conf/nips/ConneauL19}. Intuitively, parallel data contain more information than monolingual data. However, we have not observed significant improvements over pre-training with only monolingual data and the MLM objective. The performance of CINO may be improved if parallel data can be effectively used.

\textbf{Languages from different cultures}. Among the languages in Table \ref{minority_languages}, some are cross-border languages. The cross-border languages are spoken in more than one country and are influenced by local cultures. How well does the model that has been trained on the corpus collected in one country transfer to the corpus collected in another country? If the writing systems of the language are different (for example, Mongolian is written in Cyrillic in Mongolia, while it is written in traditional Mongolian script in China), to what extent do writing systems influence the model performance? We expect future work to address these questions.

\section{Conclusion}
In this paper, we introduce CINO, a multilingual pre-trained language model for Chinese minority languages. It takes the same structure as XLM-R but with a different vocabulary and is pre-trained with an adapted MLM objective to reduce computational costs. We build multilingual text classification datasets WCM from Wikipedia and CMNews from ethnic minority news for zero-shot ability evaluation on the Chinese minority languages. We evaluate CINO on several text classification tasks. The results show that CINO achieves notable improvements over the existing baselines.

\section*{Acknowledgments}
We would like to thank all anonymous reviewers for their thorough review and for providing constructive comments to improve our paper.

\bibliographystyle{acl_natbib}
\bibliography{CINO}

\clearpage
\newpage

\appendix

\section{Statistics of the Pre-training Corpora}\label{sec:stat-corpora}
The corpus size and mean sequence length  for pre-training are listed in Table \ref{corpus-statistics}. The sequence lengths are obtained by counting the tokens after tokenization. For Standard Chinese (zh), we concatenate or truncate each example to the max sequence length, while for other languages, we do not concatenate the examples but keep them unchanged.

\begin{table}[thbp]
\center
\begin{tabular}{@{}lrr@{}}
  \toprule
\textbf{Language} &  \textbf{\# Tokens} & \textbf{Mean Sequence Length} \\
\midrule
bo & 130M & 13.4      \\
kk & 238M & 60.7 \\ 
ko & 170M & 20.0 \\
mn & 337M & 25.7 \\
ug & 1B& 23.1 \\
yue & 276M & 12.6 \\
za & 23M & 58.1 \\
zh & 1.2B & 254 \\
\bottomrule
\end{tabular}
\caption{Corpus size and mean sequence length of each language in the pre-training data.}
\label{corpus-statistics}
\end{table}

\section{Hyperparameters}
\subsection{Pre-training Hyperparameters}\label{sec:pt-hp}

\begin{table}[thbp]
\center
\begin{tabular}{@{}lrr@{}}
  \toprule
\textbf{Hyperparameter} & \textbf{Base Model} & \textbf{Large Model} \\
\midrule
Batch Size    & 4,096		& 8,192        \\
Warmup Steps & 10k	& 5k	\\
Training Steps & 150k & 75k \\
Peak Learning Rate & 2e-4 & 1e-4 \\ 
Max Length & 256 & 256 \\
MLM probability & 0.2 & 0.2 \\
Adam $\epsilon$ & 1e-8 & 1e-8 \\
Adam $\beta_1$ &0.9 & 0.9 \\
Adam $\beta_2$ &0.999 & 0.999 \\
Gradient Clipping &1.0 & 1.0 \\ 
Weight Decay & 0 & 0\\ 
Sampling $\alpha$      & 0.7                             & 0.7         \\
\bottomrule
\end{tabular}
\caption{Hyperparameters used for pretraining CINO models.}
\label{hyperparameters}
\end{table}

Table \ref{hyperparameters} presents the full set of the hyperparameters used for pre-training CINO models.

\begin{table}[htbp]
  \center
  \begin{tabular}{@{}lcccc@{}}
  \toprule
  \textbf{Dataset} & \textbf{\# Train} & \textbf{\# Dev} & \textbf{\# Test} & \textbf{\# Classes} \\ \midrule
  TNCC             & 7,359           & 191           & 923   & 12          \\
  YNAT             & 45,678          & 9,106         & -    &   7   \\ \bottomrule
  \end{tabular}
  \caption{Number of examples in TNCC and YNAT.}\label{TNCC_YNAT}
  \end{table}

\begin{table*}[]
\center
\begin{tabular}{@{}l|cc|cc|cc|cc@{}}
\toprule
\multicolumn{1}{c|}{\multirow{2}{*}{\textbf{Model}}} & \multicolumn{2}{c|}{\textbf{TNCC}} & \multicolumn{2}{c|}{\textbf{YNAT}} & \multicolumn{2}{c|}{\textbf{WCM}} & \multicolumn{2}{c}{\textbf{CMNews}} \\ \cmidrule(l){2-9} 
\multicolumn{1}{c|}{}                       & \textbf{LR}         & \textbf{Epochs}       &\textbf{LR}       & \textbf{Epochs}     & \textbf{LR}        & \textbf{Epochs}     & \textbf{LR}        & \textbf{Epochs}     \\ \midrule
Word2vec (Tibetan)                           & 3e-2       & 20           & -     & -        & -          & -           & -           & -            \\
TextCNN                                       & 1e-4       & 40           & -     & -        & -          & -           & -           & -            \\
mBERT                                       & 3e-5       & 40           & 2e-5       & 5            & -          & -           & -           & -            \\
KLUE-BERT-base                              & -          & -            & 3e-5       & 3            & -          & -           & -           & -            \\
XLM-R-base                                  & 5e-5       & 40           & 3e-5       & 3            & 1e-5       & 20          & 3e-5        & 5            \\
CINO-base                                   & 5e-5       & 40           & 3e-5       & 3            & 1e-5       & 20          & 3e-5        & 5            \\
XLM-R-large                                 & 3e-5       & 40           & 2e-5       & 3            & 1e-5       & 20          & 3e-5        & 5            \\
CINO-large                                  & 3e-5       & 40           & 2e-5       & 3            & 1e-5       & 20          & 3e-5        & 5            \\ \bottomrule
\end{tabular}
\caption{Hyperparameters used for downstream fine-tuning.}
\label{ft-hyperparameters}
\end{table*}

\subsection{Fine-tuning Hyperparameters}\label{sec:ft-hp}

The hyperparameters for fine-tuning on the downstream tasks is listed in Table \ref{ft-hyperparameters}. The batch size is 32 for all experiments except Word2vec (Tibetan), of which batch size is 16. The learning rate is scheduled with 10\% warmup steps followed by a linear decay. 

We use Gensim \cite{rehurek_lrec} to train the Word2vec embeddings, and set $\texttt{min\_count}=1$, $\texttt{vector\_size}=300$. Other parameters take the default values.

\section{Statistics of the Datasets}\label{sec:stat-data}

The sizes of TNCC and YNAT are shown in Table \ref{TNCC_YNAT}.
Detailed data distribution of WCM is listed in Table \ref{WCM_distribution}.
Detailed data distribution of CMNews is listed in Table \ref{CMNews_distribution}.

\begin{table*}[htbp]
  \center
  \small
    \begin{tabular}{@{}lccccccccc@{}}
    \toprule
    \textbf{Category}      & \multicolumn{1}{c}{\textbf{mn}} & \multicolumn{1}{c}{\textbf{bo}} & \multicolumn{1}{c}{\textbf{ug}} & \multicolumn{1}{c}{\textbf{kk}} & \multicolumn{1}{c}{\textbf{ko}} & \multicolumn{1}{c}{\textbf{yue}} & \multicolumn{1}{c}{\textbf{zh-train}} & \multicolumn{1}{c}{\textbf{zh-test}} & \multicolumn{1}{c}{\textbf{zh-dev}} \\ \midrule
    \textbf{Arts}            & 135                             & 141                             & 3                               & 348                             & 806                             & 387                              & 2657                                  & 335                                 & 331                                  \\
    \textbf{Geography}       & 76                              & 339                             & 256                             & 572                             & 1197                            & 1550                             & 12854                                 & 1644                                & 1589                                 \\
    \textbf{History}         & 66                              & 111                             & 0                               & 491                             & 776                             & 499                              & 1771                                  & 248                                 & 227                                  \\
    \textbf{Nature}          & 7                               & 0                               & 7                               & 361                             & 442                             & 606                              & 1105                                  & 110                                 & 134                                  \\
    \textbf{Natural Science} & 779                             & 133                             & 20                              & 880                             & 532                             & 336                              & 2314                                  & 287                                 & 317                                  \\
    \textbf{Personage}       & 1402                            & 111                             & 0                               & 169                             & 684                             & 1230                             & 7706                                  & 924                                 & 953                                  \\
    \textbf{Technology}      & 191                             & 163                             & 8                               & 515                             & 808                             & 329                              & 1184                                  & 152                                 & 134                                  \\
    \textbf{Education}       & 6                               & 1                               & 0                               & 1392                            & 439                             & 289                              & 936                                   & 118                                 & 130                                  \\
    \textbf{Economy}         & 205                             & 0                               & 0                               & 637                             & 575                             & 445                              & 922                                   & 109                                 & 113                                  \\
    \textbf{Health}          & 106                             & 111                             & 6                               & 893                             & 299                             & 272                              & 551                                   & 73                                  & 67                                   \\ \midrule
    \textbf{Total}           & 2973                            & 1110                            & 300                             & 6258                            & 6558                            & 5943                             & 32000                                 & 4000                                & 3995                                \\ \bottomrule
    \end{tabular}
    \caption{Number of examples in each category and language in WCM.}
    \label{WCM_distribution}
\end{table*}

\begin{table*}[htbp]
  \center
  \small
\begin{tabular}{@{}c|lccccccc@{}}
\toprule
                                 \textbf{Split}           & \textbf{Category}         & \textbf{bo}              & \textbf{kk}             & \textbf{ko}              & \textbf{mn}              & \textbf{ug}              & \textbf{yue}             & \textbf{zh}              \\ \midrule
\multicolumn{1}{c|}{\multirow{9}{*}{\textbf{Train}}} & \textbf{Education} & 626                      & 364                     & 378                      & 187                      & 423                      & 880                      & 1979                     \\
\multicolumn{1}{c|}{}                       & \textbf{Sports}    & 66                       & 133                     & 321                      & 556                      & 1216                     & 70                       & 1978                     \\
\multicolumn{1}{c|}{}                       & \textbf{Health}    & 1309                     & 153                     & 40                       & 31                       & 240                      & 1358                     & 2000                     \\
\multicolumn{1}{c|}{}                       & \textbf{Tourism}   & 1128                     & 12                      & 43                       & 102                      & 1078                     & 0                        & 1998                     \\
\multicolumn{1}{c|}{}                       & \textbf{Legal}     & 433                      & 283                     & 283                      & 294                      & 19                       & 22                       & 2000                     \\
\multicolumn{1}{c|}{}                       & \textbf{Economy}   & 399                      & 107                     & 192                      & 510                      & 0                        & 1080                     & 1877                     \\
\multicolumn{1}{c|}{}                       & \textbf{Culture}   & 1834                     & 231                     & 228                      & 118                      & 0                        & 0                        & 1995                     \\
\multicolumn{1}{c|}{}                       & \textbf{Society}    & 898                      & 149                     & 147                      & 543                      & 1132                     & 169                      & 1935                     \\ \cmidrule(l){2-9} 
\multicolumn{1}{c|}{}                       & \textbf{Total}     & 6693                     & 1432                    & 1632                     & 2341                     & 4108                     & 3579                     & 15762                    \\ \midrule
\multirow{9}{*}{\textbf{Dev}}                        & \textbf{Education} & \multicolumn{1}{c}{418}  & \multicolumn{1}{c}{243} & \multicolumn{1}{c}{253}  & \multicolumn{1}{c}{125}  & \multicolumn{1}{c}{282}  & \multicolumn{1}{c}{587}  & \multicolumn{1}{c}{1000} \\
                                            & \textbf{Sports}    & \multicolumn{1}{c}{44}   & \multicolumn{1}{c}{89}  & \multicolumn{1}{c}{215}  & \multicolumn{1}{c}{371}  & \multicolumn{1}{c}{811}  & \multicolumn{1}{c}{48}   & \multicolumn{1}{c}{1000} \\
                                            & \textbf{Health}    & \multicolumn{1}{c}{874}  & \multicolumn{1}{c}{103} & \multicolumn{1}{c}{28}   & \multicolumn{1}{c}{21}   & \multicolumn{1}{c}{160}  & \multicolumn{1}{c}{906}  & \multicolumn{1}{c}{1000} \\
                                            & \textbf{Tourism}   & \multicolumn{1}{c}{752}  & \multicolumn{1}{c}{8}   & \multicolumn{1}{c}{30}   & \multicolumn{1}{c}{68}   & \multicolumn{1}{c}{719}  & \multicolumn{1}{c}{0}    & \multicolumn{1}{c}{1000} \\
                                            & \textbf{Legal}     & \multicolumn{1}{c}{289}  & \multicolumn{1}{c}{190} & \multicolumn{1}{c}{189}  & \multicolumn{1}{c}{196}  & \multicolumn{1}{c}{14}   & \multicolumn{1}{c}{15}   & \multicolumn{1}{c}{1000} \\
                                            & \textbf{Economy}   & \multicolumn{1}{c}{266}  & \multicolumn{1}{c}{72}  & \multicolumn{1}{c}{129}  & \multicolumn{1}{c}{341}  & \multicolumn{1}{c}{0}    & \multicolumn{1}{c}{721}  & \multicolumn{1}{c}{1000} \\
                                            & \textbf{Culture}   & \multicolumn{1}{c}{1223} & \multicolumn{1}{c}{155} & \multicolumn{1}{c}{152}  & \multicolumn{1}{c}{80}   & \multicolumn{1}{c}{0}    & \multicolumn{1}{c}{0}    & \multicolumn{1}{c}{1000} \\
                                            & \textbf{Society}    & \multicolumn{1}{c}{600}  & \multicolumn{1}{c}{100} & \multicolumn{1}{c}{99}   & \multicolumn{1}{c}{362}  & \multicolumn{1}{c}{756}  & \multicolumn{1}{c}{113}  & \multicolumn{1}{c}{1000} \\ \cmidrule(l){2-9} 
                                            & \textbf{Total}     & \multicolumn{1}{c}{4466} & \multicolumn{1}{c}{960} & \multicolumn{1}{c}{1095} & \multicolumn{1}{c}{1564} & \multicolumn{1}{c}{2742} & \multicolumn{1}{c}{2390} & \multicolumn{1}{c}{8000} \\ \bottomrule
\end{tabular}
    \caption{Number of examples in each category and language in CMNews.}
    \label{CMNews_distribution}
\end{table*}

\end{document}